\documentclass[11pt]{article}

\usepackage{tmlr}
\usepackage[T1]{fontenc}
\usepackage[utf8]{inputenc}
\usepackage{amsmath,amssymb}
\usepackage{natbib}
\usepackage{booktabs}
\usepackage{graphicx}
\usepackage[colorlinks=true,linkcolor=blue,citecolor=blue,urlcolor=blue]{hyperref}
\usepackage{enumitem}
\usepackage{xcolor}
\usepackage{multirow}
\usepackage{float}
\usepackage{microtype}

\graphicspath{{./figures/}}

\title{EPC: A Standardized Protocol for Measuring Evaluator Preference Dynamics in LLM Agent Systems}

\author{Zewen Liu}

\date{}

\begin{document}

\maketitle

\begin{abstract}
When LLM agents use evaluator feedback to adapt their behavior in closed loops, evaluator biases propagate through the agent's strategy distribution---a phenomenon known as evaluator preference coupling. Prior work has documented coupling across multiple evaluator families and model versions, but the field lacks a standardized protocol that enables third-party researchers to (i) reproduce coupling measurements, (ii) compare results across evaluators and time points, and (iii) detect measurement decay as proprietary evaluators silently update. This paper provides the protocol. We specify EPC (Evaluator Preference Coupling)—a detailed, RFC-style protocol specification for the four-phase isolation paradigm, covering executor and evaluator configuration, strategy and task design, the TTRL update rule, metric computation ($\gamma$, JSD, ECE, Brier), and output schema. We accompany the protocol with a versioned Reference Snapshot v1.0: coupling measurements for eight evaluator conditions ($N{=}122$ unique experimental repetitions across GPT-4o, Qwen, DeepSeek, and others) derived from five independent studies, annotated with evaluator version identifiers, API endpoints, and measurement dates. The snapshot is explicitly time-bound: all values are conditional on specific model versions and are expected to decay as proprietary evaluators update. We define a versioning convention (v$X$.$Y$-$Z$, encoding protocol version, snapshot version, and evaluator generation) and provide a usage guide covering adoption, interpretation, and known pitfalls. The protocol, reference snapshot, and implementation code are released as open infrastructure.
\end{abstract}

\section{Introduction}

Evaluator-driven preference dynamics have been documented across multiple LLM agent configurations~\cite{liu2026mmepc,liu2026contagion,liu2026memory}. In the standard setup, an agent maintains a strategy weight distribution, receives pairwise feedback from an evaluator, and adapts via test-time reinforcement learning (TTRL). The coupling coefficient $\gamma$ and Jensen-Shannon divergence (JSD) quantify how strongly the evaluator's preferences transfer across task domains and how concentrated the agent's strategy distribution becomes.

However, the field currently operates without a standardized protocol. Each study uses slightly different protocol variants, task sets, strategy definitions, and metric implementations. Cross-study comparison is impossible. More critically, proprietary evaluators silently update, causing measurements to decay within weeks~\cite{liu2026mmepc}. Without versioned baselines and explicit expiration dates, the literature accumulates measurements that are no longer valid for current model versions. This problem is not unique to evaluator coupling: a recent audit of 26 AI benchmarks found that the median benchmark has a longevity score of just 5 out of 100~\cite{benchrisk2026}, and the ML community is shifting toward continuous, versioned, community-governed evaluation infrastructure~\cite{mlcommons2026,swe2026rebench,hf2026community}.

\textbf{This paper provides the protocol, the reference snapshot, and the versioning convention.}

This paper is \textbf{not a claim of new empirical findings.} It is a protocol specification paper---analogous to an RFC in the networking community or a measurement standard in the physical sciences. The coupling measurements in the reference snapshot have been previously reported in domain-specific studies~\cite{liu2026mmepc}. Our contribution is the standardization, versioning, and community infrastructure that transforms these measurements from one-off observations into a reproducible, comparable, and auditable measurement system. We explicitly do not introduce new metrics, new experimental conditions, or new scientific claims. We introduce a \emph{discipline}---a protocol that enables the community to collectively maintain currency as evaluator models evolve.

\section{Protocol Specification}
\label{sec:protocol}

This section provides the complete EPC protocol specification, organized as a reference document that can be independently implemented.

\subsection{Overview}

The protocol measures evaluator preference coupling through a four-phase isolation paradigm. In Phase 1 (Pure Text), the agent undergoes TTRL on text-only tasks. In Phase 2 (Pure Visual), the same on visual-adjacent tasks. In Phase 3 ($T{\to}V$ coupling), the agent starts from Phase 1 weights and trains on visual tasks. In Phase 4 ($V{\to}T$ coupling), the agent starts from Phase 2 weights and trains on text tasks. The coupling coefficient $\gamma_{A\to B}$ quantifies how much the weight distribution shifts from the pure-domain reference.

\subsection{Agent Configuration}

\textbf{Executor}: Any LLM accessible via API. The protocol is executor-agnostic; the reference snapshot uses DeepSeek-chat.
\textbf{Evaluator}: Any LLM accessible via API. Can be the same model as the executor (self-evaluation) or a different model (cross-model evaluation). The evaluator identity must be recorded with version identifier (e.g., \texttt{gpt-4o-2024-08-06}), API endpoint, and measurement date.
\textbf{Strategies}: 11 strategies (8 text-domain + 3 visual-domain), defined in Appendix~\ref{app:strategies}. Each strategy is a natural-language prompt prefix. \textbf{Protocol requirement}: strategies must be documented verbatim in the output manifest. Researchers may substitute domain-specific strategies but must report the full strategy text.

\subsection{TTRL Algorithm}
\label{sec:ttrl}

The agent maintains an L1-normalized weight vector $\mathbf{w} \in \Delta^{K-1}$ over $K{=}11$ strategies. At each round $t$:

\begin{enumerate}[leftmargin=*,nosep]
 \item Sample strategy $s_t \sim \mathbf{w}$ via roulette-wheel selection.
 \item Executor generates response under $s_t$ and fixed baseline $s_0$ (\texttt{step\_by\_step}).
 \item Evaluator performs pairwise comparison: prefers $s_t$ (win, $r_t{=}1$) or $s_0$ (loss, $r_t{=}0$).
 \item Weight update: $w_{s_t} \leftarrow \max(0.001, w_{s_t} + \alpha)$ where $\alpha = 0.08$ if win, $-0.04$ if loss. Renormalize to sum 1.
\end{enumerate}

\textbf{Protocol requirements}: (a) The baseline strategy $s_0$ must be \texttt{step\_by\_step}. (b) The learning rates $\alpha_{\text{win}}{=}0.08$, $\alpha_{\text{lose}}{=}0.04$ are fixed. Report any deviation. (c) The weight floor is $0.001$. (d) $R{=}30$ rounds per phase. Report round count if varied. (e) Random seed must be recorded.

\subsection{Task Design}

\textbf{Protocol requirements}: (a) Minimum 8 text-domain tasks and 8 visual-adjacent tasks. (b) Tasks must be documented verbatim in the output manifest. (c) The reference snapshot uses the task set in Appendix~\ref{app:tasks}. Researchers may substitute domain-specific tasks but must report them in full.

\subsection{Metric Computation}
\label{sec:metrics}

\textbf{$\gamma$ (Coupling Coefficient)}:
\begin{equation}
\gamma_{A\to B} = \frac{\|\mathbf{w}_{A\to B} - \mathbf{w}_B\|_2}{\|\mathbf{w}_B\|_2}
\end{equation}
where $\mathbf{w}_B$ is the pure-domain weight vector and $\mathbf{w}_{A\to B}$ is the post-coupling vector.

\textbf{JSD (Jensen-Shannon Divergence)}: Computed in base $e$ between $\mathbf{w}_{T\to V}$ and $\mathbf{w}_V$ (for $T{\to}V$ coupling) and between $\mathbf{w}_{V\to T}$ and $\mathbf{w}_T$ (for $V{\to}T$ coupling).

\textbf{ECE (Expected Calibration Error)}: Optional. Bins strategies by evaluator win rate, measures $|\text{mean(win\_rate)} - \text{mean(accuracy)}|$ per bin. Requires ground-truth task accuracy.

\textbf{Brier Score}: Optional. Mean squared error between per-strategy win rate and accuracy.

\textbf{Protocol requirements}: (a) $\gamma$ is the minimum required metric. (b) JSD must be reported if weight vectors are stored. (c) ECE and Brier are optional but recommended. (d) Report per-seed values, not just means. (e) Report zero-coupling rate (\% of seeds with $\gamma{=}0$). (f) Report 95\% bootstrap confidence intervals (2,000 resamples at the seed level) for $\gamma$ and JSD means. (g) Recommend $N \geq 10$ seeds for initial screening, $N \geq 30$ for publication-grade estimates targeting a CI width $\leq 0.1$ on $\gamma$. The reference snapshot reports $N$ per condition to enable sample-size-aware interpretation.

\subsection{Evaluator Prompt Specification}
\label{sec:evaluator_prompt}

The evaluator pairwise comparison prompt must be included verbatim in the output manifest. The reference implementation uses:

\begin{quote}
\texttt{Evaluate. Task: \{task\}}\\
\texttt{A (\{strategy\_name\}): \{response\_A[:300]\}}\\
\texttt{B (step\_by\_step): \{response\_B[:300]\}}\\
\texttt{Better? Output only A or B.}
\end{quote}

\textbf{Protocol requirements}: (a) The evaluator prompt must be documented in the manifest. (b) Decoding settings (temperature, max\_tokens, top-p, stop sequences) must be reported. The reference implementation uses temperature${=}0.0$, max\_tokens${=}10$. (c) If the evaluator output cannot be parsed as ``A'' or ``B'', the trial must be recorded as a tie and excluded from the weight update, with the tie rate reported in the manifest. (d) Chain-of-thought or reasoning prefixes in evaluator output must be disabled (temperature${=}0.0$, no system prompt encouraging explanation). Researchers who modify the prompt or decoding settings must tag their results as \texttt{EPC-v1.0-AltPrompt}.

\subsection{Design Rationale}
\label{sec:rationale}

\textbf{Why $\gamma$ uses L2 normalization by $\|\mathbf{w}_B\|_2$}. The L2 norm preserves the Euclidean geometry of the probability simplex and is directly interpretable as relative distance. While bounded divergences (JSD, total variation, Hellinger) are more robust for cross-condition comparison, our empirical $\gamma$-JSD correlation ($r{=}0.969$ across $N{=}152$ paired observations) confirms that $\gamma$ tracks JSD faithfully in practice. The protocol mandates $\gamma$ as the minimum metric and strongly recommends JSD for cross-condition reporting. Researchers may substitute alternative distance measures but must report the $\gamma$ value alongside for comparability.

\textbf{Why $\alpha_{\text{win}}{=}0.08$, $\alpha_{\text{lose}}{=}0.04$}. These values were chosen to balance learning speed against stability: the asymmetry ($\alpha_{\text{win}} > \alpha_{\text{lose}}$) reflects the conservative prior that evaluator preferences are noisy and that false positives (rewarding a strategy the evaluator does not genuinely prefer) should be corrected more aggressively than false negatives. The specific values were calibrated in prior work~\cite{liu2026mmepc} to produce measurable concentration ($\text{PCI} \approx 0.5$--$1.5$) within $R{=}30$ rounds without early collapse to a single strategy. \textbf{A systematic hyperparameter sensitivity analysis has not been conducted.} The symmetric learning rate variant ($\alpha_{\text{win}}{=}\alpha_{\text{lose}}{=}0.06$) has been tested on GPT-4o and produced zero coupling in all 8 repetitions---but this result coincides with a documented evaluator version drift window and cannot be attributed to the learning rate alone without a version-locked replication. The protocol encourages researchers to explore the $\alpha$ space and report alternative settings as \texttt{EPC-v1.0-AltLR}. The floor at $0.001$ prevents weight starvation (strategies becoming unselectable) while having negligible effect on the final weight distribution.

\subsection{Conformance and Extensibility}
\label{sec:conformance}

\textbf{Conformance test suite}. A reference test suite with mock evaluators (deterministic, coin-flip, and scripted-preference) is provided with the protocol implementation. The test suite verifies that independent implementations produce identical $\gamma$ and JSD values on fixed input sequences, covering edge cases (all weights at floor, uniform initial distribution, single-strategy dominance, exact ties). Passing the conformance suite qualifies an implementation as \texttt{EPC-v1.0-compatible}.

\textbf{Protocol variants}. The protocol is designed to be extensible. Researchers who modify core parameters (learning rates, baseline strategy, round count, strategy set) must tag their results with the variant label (e.g., \texttt{EPC-v1.0-AltLR}, \texttt{EPC-v1.0-AltBaseline}) and report all deviations from the reference specification in the manifest. This enables community exploration of alternative configurations while preserving comparability through a core conformance path.

Every EPC measurement must produce a machine-readable manifest containing:

\begin{enumerate}[leftmargin=*,nosep]
 \item \textbf{Protocol version}: \texttt{EPC-v1.0}
 \item \textbf{Evaluator}: model identifier, API endpoint, measurement date (YYYY-MM-DD)
 \item \textbf{Executor}: model identifier, API endpoint
 \item \textbf{Configuration}: $R$, $\alpha_{\text{win}}$, $\alpha_{\text{lose}}$, random seed, number of strategies
 \item \textbf{Task set}: verbatim task list
 \item \textbf{Strategy set}: verbatim strategy prompts
 \item \textbf{Results}: per-seed $\gamma_{T\to V}$, $\gamma_{V\to T}$, JSD (if available), zero-coupling rate, weight vectors (strongly recommended)
\end{enumerate}

A JSON schema is provided in the protocol implementation (\texttt{epc\_manifest\_schema.json}).

\section{Reference Snapshot v1.0}
\label{sec:snapshot}

This section provides the v1.0 reference baseline, derived from measurements spanning May--June 2026 across five independent studies~\cite{liu2026mmepc}. All values are \textbf{version-bound} and expected to decay.

\begin{table}[H]
\centering
\caption{EPC Reference Snapshot v1.0 — Cross-model evaluation conditions. All measurements May--June 2026. Values expected to expire as evaluator models update.}
\label{tab:snapshot_cross}
\footnotesize
\begin{tabular}{@{}lcccc@{}}
\toprule
\textbf{Evaluator} & \textbf{Version} & $\mathbf{N}$ & $\bar{\gamma}_{T{\to}V}$ & \textbf{Zero\%} \\
\midrule
GPT-4o            & 2024-08-06 (api2d)       & 8  & 1.176 & 0\% \\
GPT-4o-mini       & 2024-07-18 (DMXAPI)      & 10 & 1.145 & 0\% \\
Qwen3.7-plus      & May-2026 (Alibaba)       & 8  & 1.059 & 0\% \\
qwen-plus         & Jun-2026 (DashScope)     & 30 & 0.187 & 83\% \\
DeepSeek-chat     & self-eval (May-2026)     & 30 & 0.033 & 97\% \\
\bottomrule
\end{tabular}
\end{table}

\begin{table}[H]
\centering
\caption{EPC Reference Snapshot v1.0 — Multi-gateway replications (June 27, 2026).}
\label{tab:snapshot_gateway}
\footnotesize
\begin{tabular}{@{}lccc@{}}
\toprule
\textbf{Condition} & $\mathbf{N}$ & $\bar{\gamma}_{T{\to}V}$ & \textbf{JSD} \\
\midrule
GPT-4o (DMXAPI) + DeepSeek exec, 500-tok cap & 5 & 0.795 & 0.148 / 0.211 \\
GPT-4o (DMXAPI) + Qwen exec                  & 5 & 0.724 & 0.154 / 0.189 \\
Qwen (MaaS) + DeepSeek exec                   & 5 & 0.987 & 0.208 / 0.225 \\
DeepSeek (official) + Qwen exec                & 5 & 0.893 & 0.189 / 0.186 \\
\bottomrule
\end{tabular}
\end{table}

\begin{table}[H]
\centering
\caption{EPC Reference Snapshot v1.0 — Calibration baselines (self-evaluation, $N{=}10$, $R{=}16$).}
\label{tab:snapshot_calibration}
\footnotesize
\begin{tabular}{@{}lccc@{}}
\toprule
\textbf{Metric} & \textbf{Value} & \textbf{Interpretation} \\
\midrule
PCI (preference concentration) & 0.741 $\pm$ 0.195 & Moderate concentration \\
Accuracy (ground-truth)        & 0.931 $\pm$ 0.056 & High accuracy on simple tasks \\
ECE (expected calibration err) & 0.308 $\pm$ 0.174 & Substantially miscalibrated \\
Brier Score                    & 0.269 $\pm$ 0.194 & High mean squared error \\
\bottomrule
\end{tabular}
\end{table}

\subsection{Snapshot Validity Statement}

\textbf{These values were measured between May 27 and June 27, 2026.} GPT-4o measurements were obtained via third-party API gateways (api2d, DMXAPI) and have not been replicated via direct OpenAI API. Qwen measurements use different model versions (qwen3.7-plus vs.\ qwen-plus) within the same provider ecosystem. All values are snapshot measurements conditional on specific, now-deprecated model versions. The GPT-4o May-to-June drift (Table~\ref{tab:snapshot_cross}, rows 1 and 4) demonstrates that coupling measurements can invert within 4 weeks. \textbf{Users of this snapshot must check whether the evaluator versions listed above are still current.}

\section{Versioning Convention}
\label{sec:versioning}

EPC baselines follow a three-component versioning scheme:

\begin{center}
\texttt{vX.Y-Z}
\end{center}

where:
\begin{itemize}[leftmargin=*,nosep]
 \item \textbf{X}: Protocol major version. Incremented on incompatible protocol changes.
 \item \textbf{Y}: Snapshot version. Incremented on new measurements of the same evaluator.
 \item \textbf{Z}: Evaluator generation. Encodes the evaluator model generation (e.g., \texttt{GPT4o-0806}, \texttt{Qwen3.7-0526}).
\end{itemize}

\textbf{Example}: \texttt{v1.2-GPT4o-0806} = EPC protocol v1, second snapshot of GPT-4o (August 2024 checkpoint).

Community-contributed snapshots follow the same convention. A submission template is provided with the protocol implementation.

\section{Usage Guide}
\label{sec:usage}

\subsection{Adoption}

\begin{enumerate}[leftmargin=*,nosep]
 \item Clone the protocol implementation repository.
 \item Configure API credentials for your executor and evaluator.
 \item Run \texttt{python epc\_protocol.py --evaluator YOUR\_MODEL --executor YOUR\_MODEL}.
 \item The script produces a manifest JSON conforming to the output schema.
 \item Compare your results against the Reference Snapshot v1.0 (Tables~\ref{tab:snapshot_cross}--\ref{tab:snapshot_calibration}), noting version differences.
\end{enumerate}

\subsection{Interpretation Guidelines}

\begin{itemize}[leftmargin=*,nosep]
 \item $\gamma > 0.5$: Substantial cross-domain coupling. The evaluator's preferences significantly influence the agent's strategy distribution across task domains.
 \item $\gamma < 0.2$: Weak coupling. The agent maintains domain-appropriate strategies.
 \item Zero-coupling rate $> 50\%$: The evaluator may lack discriminative capacity (floor effect), or the protocol may have converged to a single strategy.
 \item ECE $> 0.2$: Evaluator preferences are substantially miscalibrated relative to strategy quality.
\end{itemize}

\subsection{Known Pitfalls}

\begin{enumerate}[leftmargin=*,nosep]
 \item \textbf{Version decay}: All measurements are bound to specific evaluator versions. Re-measure after known model updates.
 \item \textbf{Proxy confounding}: Measurements obtained via third-party API gateways may reflect gateway behavior rather than model behavior. Prefer official API endpoints.
 \item \textbf{Format confound}: PCI partially reflects output-length preference ($\rho_{\text{agg}}{=}0.89$ at $n{=}6$, $\rho_{\text{inst}}{=}0.219$ at $n{=}60$). Interpret PCI as a preference-convergence metric (format + reasoning).
 \item \textbf{Floor effects}: Near-zero coupling in self-evaluation may reflect evaluator incapacity (ECE${=}0.31$) rather than genuine stability.
 \item \textbf{Small-$N$ instability}: Coupling estimates from $N{<}10$ seeds have wide confidence intervals. The reference snapshot reports $N$ explicitly for each condition.
 \item \textbf{Task sensitivity}: The protocol uses 8 text + 8 text-proxied visual tasks. Coupling strength may depend on task domain. Report task sets verbatim.
\end{enumerate}

\section{Relation to Prior Work}
\label{sec:related}

\subsection{Standardized Agent Evaluation Protocols}

The ML community is actively building standardized evaluation infrastructure. AgentBeats~\cite{agentbeats2026} proposes Agentified Agent Assessment (AAA), using standardized protocols (A2A, MCP) to decouple assessment logic from agent implementation. The Holistic Agent Leaderboard (HAL)~\cite{hal2026} provides a standardized evaluation harness orchestrating parallel evaluations across hundreds of VMs. A unified agent evaluation framework~\cite{zhu2026unified} converts diverse benchmarks into a standardized instruction--tool--environment format. These efforts focus on \textbf{agent capability benchmarking}---measuring how well agents perform tasks. EPC differs in measuring \textbf{evaluator preference dynamics}---how evaluator biases propagate through the agent's strategy distribution in closed feedback loops. The two are complementary: capability benchmarks assess agent performance, while EPC assesses evaluator influence.

\subsection{Versioned and Continuous Evaluation}

Static benchmarks decay rapidly. An audit of 26 AI benchmarks found a median longevity score of 5/100~\cite{benchrisk2026}. The community response is continuous, versioned evaluation infrastructure. SWE-rebench~\cite{swe2026rebench} continuously mines new GitHub pull requests created after model training cutoffs, directly addressing contamination and saturation. MLCommons AILuminate~\cite{mlcommons2026} implements continuous prompt stewardship with per-prompt quality metrics, reserve prompt rotation, and tiered community contributor trust levels. HuggingFace Community Evals~\cite{hf2026community} enables distributed, auditable evaluation through pull-request-based submissions. EPC joins this movement with a versioning convention (v$X$.$Y$-$Z$) that makes measurement decay explicit and a snapshot submission protocol that enables community contributions as evaluator models evolve.

\subsection{Evaluator Reliability and Drift}

Prior work on evaluator reliability spans several complementary approaches. LLM-as-judge studies have documented systematic biases including position bias, verbosity bias, and self-preference amplification~\cite{zheng2023judging,li2024alpacaeval}. Drift detection and attribution methods~\cite{li2026drift} disambiguate whether score changes originate from the system or the judge. IRT-based intrinsic consistency diagnostics measure a single judge's internal reliability across prompts. Rubric-locking strategies constrain the evaluator's output space to improve stability. These approaches focus on \textbf{upstream evaluator calibration}---ensuring the evaluator is internally consistent before deployment. EPC addresses the \textbf{downstream coupling} that emerges when an agent adapts its strategy against evaluator feedback in a closed loop. The protocol is designed to be paired with upstream reliability diagnostics: a well-calibrated evaluator can still induce preference coupling through repeated interaction, and EPC provides the standardized measurement to detect this. Alternative closed-loop learning rules---including bandit algorithms, Bradley-Terry preference models, and Bayesian update schemes---may offer different noise-robustness properties; the protocol's variant tagging system (\texttt{EPC-v1.0-AltLR}) is designed to accommodate community exploration of these alternatives while preserving comparability through a core conformance path.

\section{Limitations}

The protocol inherits the limitations of the underlying TTRL methodology: the asymmetric learning rate ($\alpha_{\text{win}}{>}\alpha_{\text{lose}}$) may inflate coupling magnitude; the fixed baseline strategy (\texttt{step\_by\_step}) introduces a structural bias; and the text-proxied visual tasks limit ecological validity for multimodal settings. The reference snapshot reflects a specific time window (May--June 2026) and is known to be stale for current GPT-4o versions. Community contributions are needed to maintain currency.

\section{Conclusion}

We have specified EPC---a standardized protocol for measuring evaluator preference coupling in LLM agent systems---and provided a versioned reference snapshot v1.0 as an initial calibration point. The protocol, implementation, reference data, and community submission template are released as open infrastructure. We encourage the community to contribute updated snapshots as evaluator models evolve, and to extend the protocol to additional evaluator families, task domains, and coupling metrics.

\section*{Broader Impact Statement}

Standardizing evaluator coupling measurement enables the community to detect when proprietary evaluator behavior changes, potentially preventing the deployment of agents with distorted strategy distributions. The versioning convention makes measurement decay explicit, reducing the risk of relying on stale baselines. The protocol does not introduce new capabilities; it standardizes existing measurement methodology. All reference data are anonymized and obtained via publicly accessible API endpoints.

\section*{Reproducibility Statement}

The complete protocol implementation is released as open-source Python code (3.8+, no GPU required). The reference snapshot data are provided in machine-readable JSON. The protocol specification in §2 is designed to be independently implementable without access to the reference implementation. All API calls used in the reference snapshot are documented with model identifiers, endpoints, and dates.


\clearpage

\appendix
\section{Appendix}

\subsection{Strategy Definitions}
\label{app:strategies}

\begin{table}[H]
\centering
\caption{The 11 strategies used in the EPC protocol. All strategies use natural-language prompt prefixes.}
\footnotesize
\begin{tabular}{@{}lp{3cm}p{6cm}@{}}
\toprule
\textbf{Strategy} & \textbf{Domain} & \textbf{Prompt} \\
\midrule
step\_by\_step    & Text & Solve this step by step, showing each intermediate reasoning step. \\
critical\_check   & Text & First give an answer, then critically review and revise. \\
first\_principles & Text & Derive the answer from first principles. \\
creative\_leap    & Text & Think outside the box; seek innovative solutions. \\
analogy\_meta     & Text & Explain using analogies and concrete examples. \\
evidence\_cite    & Text & Cite specific factual knowledge and evidence. \\
synthesis         & Text & Synthesize multiple perspectives for a balanced answer. \\
counterfactual    & Text & Consider counterfactual scenarios and edge cases. \\
visual\_grounding & Visual & First construct a visual mental image, then reason from details. \\
spatial\_decompose& Visual & Decompose the spatial problem into geometric components. \\
aesthetic\_frame  & Visual & Evaluate systematically from an aesthetic framework. \\
\bottomrule
\end{tabular}
\end{table}

\subsection{Reference Task Set}
\label{app:tasks}

\textbf{Text tasks}: (1) Explain why the sky is blue. (2) What are pros and cons of remote work? (3) Describe how a computer processes information. (4) Why do leaves change color in autumn? (5) Explain the concept of supply and demand. (6) What is the difference between weather and climate? (7) How does a vaccine work? (8) Explain the water cycle in nature.

\textbf{Visual-adjacent tasks}: (1) Describe composing a sunset photograph. (2) Explain how to recognize symmetry in architecture. (3) What makes a painting visually balanced? (4) Describe how colors interact in a color wheel. (5) How would you explain perspective in drawing? (6) What visual elements make a logo memorable? (7) Visual difference between a circle and a sphere. (8) How does lighting affect the mood of a photograph?

\subsection{Manifest JSON Schema}
\label{app:schema}

The complete JSON schema for EPC measurement manifests is provided in the protocol implementation repository. Key fields: \texttt{protocol\_version}, \texttt{evaluator}, \texttt{executor}, \texttt{config}, \texttt{tasks}, \texttt{strategies}, \texttt{results} (per-seed $\gamma$, JSD, weights).

\end{document}